\pdfoutput=1 
\documentclass[runningheads]{llncs}

\usepackage{eccv} 

\usepackage{eccvabbrv}

\usepackage{graphicx}
\usepackage{booktabs}

\usepackage[accsupp]{axessibility}  

\usepackage{hyperref}

\usepackage{orcidlink}

\begin{document}

\title{Evaluating Anterior Cruciate Ligament Risk of Injury Using a Graph Convolutional Autoencoder} 

\titlerunning{ACL Injury Risk Evaluation via Graph Convolutional Autoencoder}

\author{Athanasios Babouras\inst{1}\orcidlink{0000-0003-4678-3699} \and
Paul A. Martineau\inst{2} \and
Thomas Fevens\inst{3}\orcidlink{0000-0003-1160-1429}}

\authorrunning{A. Babouras et al.}

\institute{
Department of Surgical and Interventional Sciences\\
McGill University, Montreal, QC, Canada\\  
\email{athanasios.babouras@mail.mcgill.ca} \and
Division of Orthopaedic Surgery\\
McGill University, Montreal, QC, Canada\\ 
\email{paul.martineau@mcgill.ca} \and
Department of Computer Science and Software Engineering\\
Concordia University, Montreal, QC, Canada\\
\email{thomas.fevens@concordia.ca} 
}

\maketitle

\begin{abstract}
  Current solutions for assessing the risk of anterior cruciate ligament (ACL) injury typically rely on expensive motion-capture labs that require substantial resource investment or subjective evaluations of specific athletic stress tests by experts. The Drop Vertical Jump (DVJ) and the Single Leg Hop (SLH) are two examples of such tests typically used to track progress in ACL rehabilitation. In this paper, we develop an unsupervised learning model that runs entirely on a smartphone for two tasks, based on an autoencoder architecture with multiple layers of graph convolutional networks, which utilizes pose estimation data to determine a person's injury risk status. Our model accurately predicts athletes at high risk of injury for both athletic stress tests, with average sensitivities and specificities of 78\% and 78\% for the DVJ, and 75\% and 74\% for the SLH. Our findings support the development of a widely accessible smartphone solution for real-time screening of ACL injury risk.
\end{abstract}

\section{Introduction}
\label{sec:intro}

Non-contact soft tissue injuries have been a long-standing and significant concern within the sports world and the general public. In particular, ACL injuries are one of the most common knee injuries worldwide, with an estimated incidence of more than 120,000 annually in the United States alone \cite{Kaeding2017-cf}. The vast majority of these injuries are non-contact, often resulting from a sudden cutting movement that places significant stress on the ligaments \cite{Boden2022-cu}. Such an injury can be extremely painful, may lead athletes to lose an entire season of their sport, and typically requires surgical treatment. 

Researchers have made significant efforts to assess human biomechanics to identify individuals at higher risk of ACL injury in advance, enabling targeted interventions to prevent future injuries. Hewett et al.~showed that there are significant biomechanical differences between high-risk and low-risk female athletes during a drop vertical jump (DVJ), an athletic stress test often used to assess rehabilitation progress after ACL reconstruction \cite{Hewett2005-qx}. Additionally, the single-leg hop (SLH) has been used to assess differences in movement patterns after ACL reconstruction, suggesting it could serve as a simpler alternative for risk screening. \cite{Welling2018-ab}. Typically, athletic stress tests such as the DVJ and the SLH are either performed in front of an expert who will then assess them to the best of their ability, or they are captured and analyzed using state-of-the-art motion capture labs, which can be costly and time-consuming and may not be feasible when assessing an entire sports team.

In recent years, studies have shown the potential of portable, inexpensive systems for such assessments. In their 2023 paper, Corban et al.~developed a Kinect V2-based system to detect varsity athletes at high risk of ACL injury, using common biomechanical parameters such as knee valgus and knee flexion, at specific time points throughout a DVJ \cite{Corban2023-ns}. Novel pose estimation frameworks have also shown great potential in serving as an alternative to skeleton tracking from depth sensors such as the Kinect V2 \cite{Lafayette2022-yy,Babouras2024-jv}. Although more solutions have been developed and validated for human biomechanics assessment, most DVJ and SLH evaluations still rely on discrete variables extracted from the jump rather than examining whole sections of the full jump movement and the temporal relationships among the joints throughout.

Recent advancements in machine learning, ranging from graph neural networks (GNN) to transformers, have enabled many models to effectively detect spatio-temporal patterns in data \cite{pmlr-v162-lan22a,Amato2020-oy}. To our knowledge, no published model leverages temporally sequenced joint movement patterns collected during an athletic stress test to classify ACL injury risk status. In this paper, we develop an unsupervised learning autoencoder model based on an existing GNN architecture, which takes as input a sequence of joint positions from a DVJ or SLH, captured on a smartphone running a pose estimation framework, to detect movement patterns linked to ACL tears.  We report the results of our experiments, which show that our models perform a discriminating, real-time assessment of ACL injury risk while running on a common modern smartphone.

\section{Related Work}
\label{sec:related_work}

\subsection{Injury Risk Assessment}

The bulk of the literature on the assessment of the risk of ACL injury using biomechanical evaluation has focused on measuring the angles and forces of the lower extremities during key moments of specific movements \cite{Hewett2005-qx,Bates2020-tb,Butler2024-hi}. Typically, analysis is performed using a motion capture system in conjunction with force platforms, although recent studies have shown the potential for less expensive and portable setups \cite{Corban2023-ns}. Some studies have suggested combining multiple screening tests for a more accurate assessment \cite{Butler2024-hi}. Recently, Jauhiainen et al.~explored the potential to use machine learning to evaluate an extensive array of variables across a set of screening tests for risk assessment, but reported weak predictive ability \cite{Jauhiainen2022-fe}. Currently, a widely applicable method is needed to detect the risk of ACL injury, especially for younger athletes, and to enable the use of this information to create tailored training protocols that protect high-risk athletes \cite{Hewett2016-wo}.

\subsection{Spatio-Temporal Analysis}

The use of continuous movement patterns from an athletic stress test for risk assessment has not yet been widely explored, and recent advancements in computer vision and deep learning have made it possible to do so with minimal setup. Latorre et al.~showed that a Kinect V2 can be used as a low-cost alternative for gait assessment \cite{Latorre2019-wj}. This study demonstrated the ability to analyze spatiotemporal variables using a single depth sensor with skeleton tracking for use in a clinical setting. More recently, Kumar et al.~showed that Google's MediaPipe framework, combined with a Convolutional Neural Network (CNN) model, could accurately perform gesture recognition, illustrating the potential to extract spatial information in real time from a single smartphone camera \cite{Kumar2023-pj}.

In 2018, Yan et al.~introduced a model named Spatial–Temporal Graph Convolutional Network (ST-GCN) which used an architecture based on GNNs to perform action recognition using human skeleton data \cite{Yan2018-ox}. More recent papers have focused on optimizing the graph convolutional network (GCN) architecture to more efficiently encode the topological information of skeletal joints \cite{Lee2022-xw,Zhou2024-tj}. Most solutions, however, typically rely on well-defined action classes to train their networks, which does not fit well with the problem we are attempting to solve, as the differences in spatio-temporal biomechanics between high- and low-risk individuals are not well defined.

Many researchers have suggested unsupervised methods to capture action representation from skeleton sequences, using encoder-decoder architectures \cite{Gui2018-qe,Kundu2018-ws}. Recently, a pose estimation model, BlazePose, was combined with the ST-GCN model to create a real-time fall detection model that outperformed all existing algorithms \cite{Bazarevsky2020-re,Zhang2023-uu}. Taking into account all existing work, as well as the nature of our dataset, we propose an autoencoder model with multiple layers of the ST-GCN architecture in both the encoder and decoder to detect high- and low-level movement patterns associated with an increased risk of ACL injury.

\section{Methods}

\subsection{Dataset}

Demographic characteristics and pose estimation data from two athletic stress tests were collected during the 2023 and 2024 preseason physicals for multiple varsity teams at two universities. We captured a total of 537 varsity athletes, 73 of whom were captured during both years. Data were collected according to ethical guidelines and received approval from the Research Ethics Office of the Faculty of Medicine and Health Sciences of McGill University. Participant demographics are included in Table \ref{participant_demographics}.

\begin{table}[ht]
\centering
\caption{Mean ± Standard Deviation for Participant Demographics}
\begin{tabular}{lccccc}
\toprule
 & Age (years) & Height (m) & Weight (kg) & BMI (kg/m\textsuperscript{2}) \\
\midrule
Female  & \(20 \pm 2\)  & \(1.69 \pm 0.07\)  & \(66.84 \pm 8.30\) & \(23.49 \pm 2.73\) \\
Male  & \(21 \pm 2\)  & \(1.82 \pm 0.07\)  & \(86.51 \pm 15.54\) & \(25.95 \pm 4.08\) \\
\bottomrule
\end{tabular}
\label{participant_demographics}
\end{table}

The athletes performed a DVJ and a SLH. A DVJ is performed by standing on a box, typically 31 centimetres high, dropping off the box with both legs, and then immediately jumping vertically as high as possible. To facilitate the movement, participants were instructed to stand on their heels at the edge of the box and lean forward to let their center of gravity drop them off. Each participant performed this jump three times. For the SLH, we asked participants to stand on one leg and jump forward as far as possible, landing on the same leg. The participants performed the jump three times consecutively on each leg. Due to introducing the SLH at a later stage, several participants not performing either of the two jumps due to injury, and some jumps being improperly captured, our final dataset comprised of 1630 DVJ and 2157 SLH recordings. 

We monitored the athletes over the course of two seasons, recording any non-contact ACL injuries that occurred during that period. Of the entire cohort, only six athletes were injured during the follow-up time. All six were captured performing a DVJ, while only two were also captured performing a SLH. Following the exclusion of faulty recordings, a total of 18 positive DVJ and nine positive SLH recordings were retained. Interestingly, all six injured athletes were male, despite the fact that females are known to be at much higher risk of ACL injury \cite{Parsons984}. For the purposes of model training, athletes who did not sustain injuries were considered as low risk.

\begin{figure}[ht]
\centering
\begin{minipage}{0.34\textwidth}
\centering
\includegraphics[width=0.5\columnwidth]{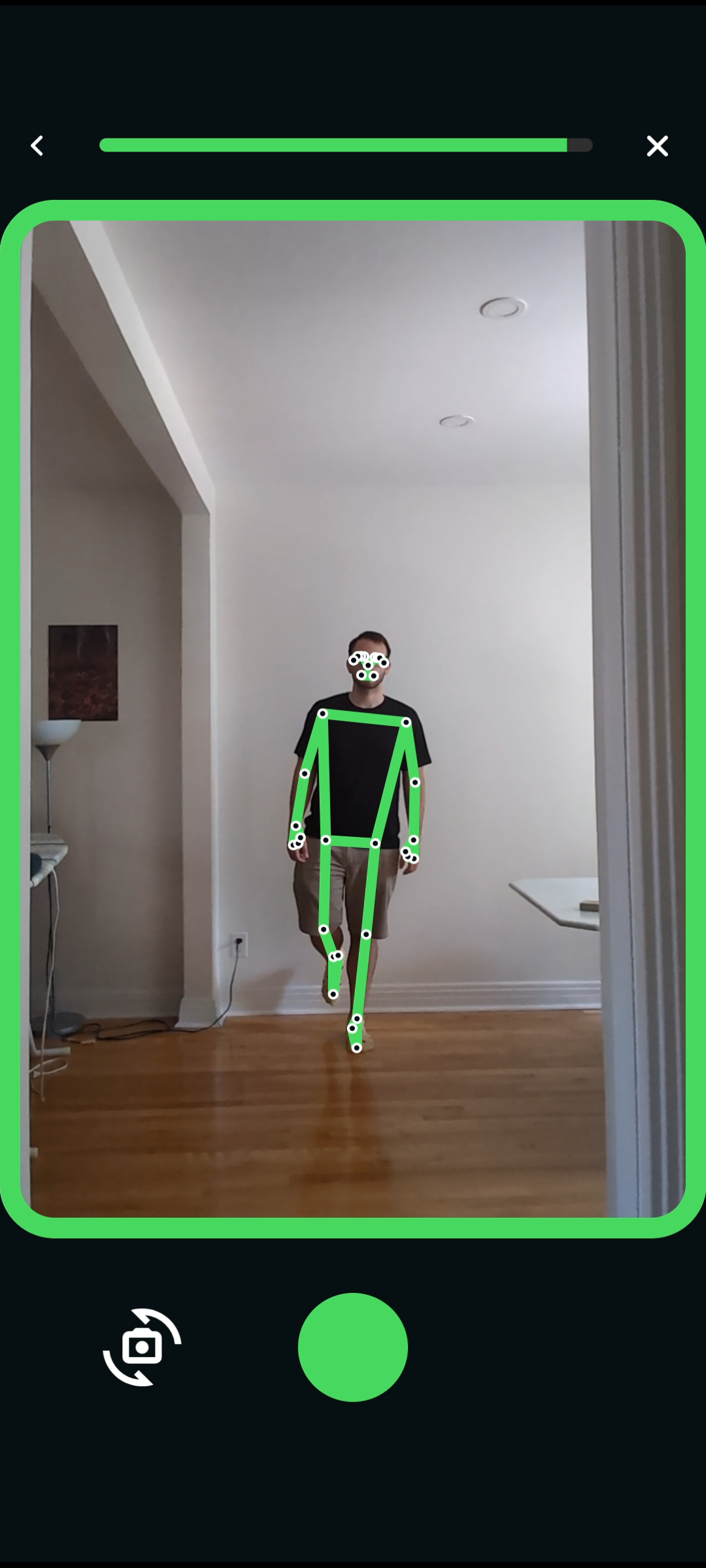} 
\caption{Smartphone Application Interface}
\label{smartphone-app}
\end{minipage}\hfill
\begin{minipage}{0.64\textwidth}
\centering
\includegraphics[width=0.9\columnwidth]{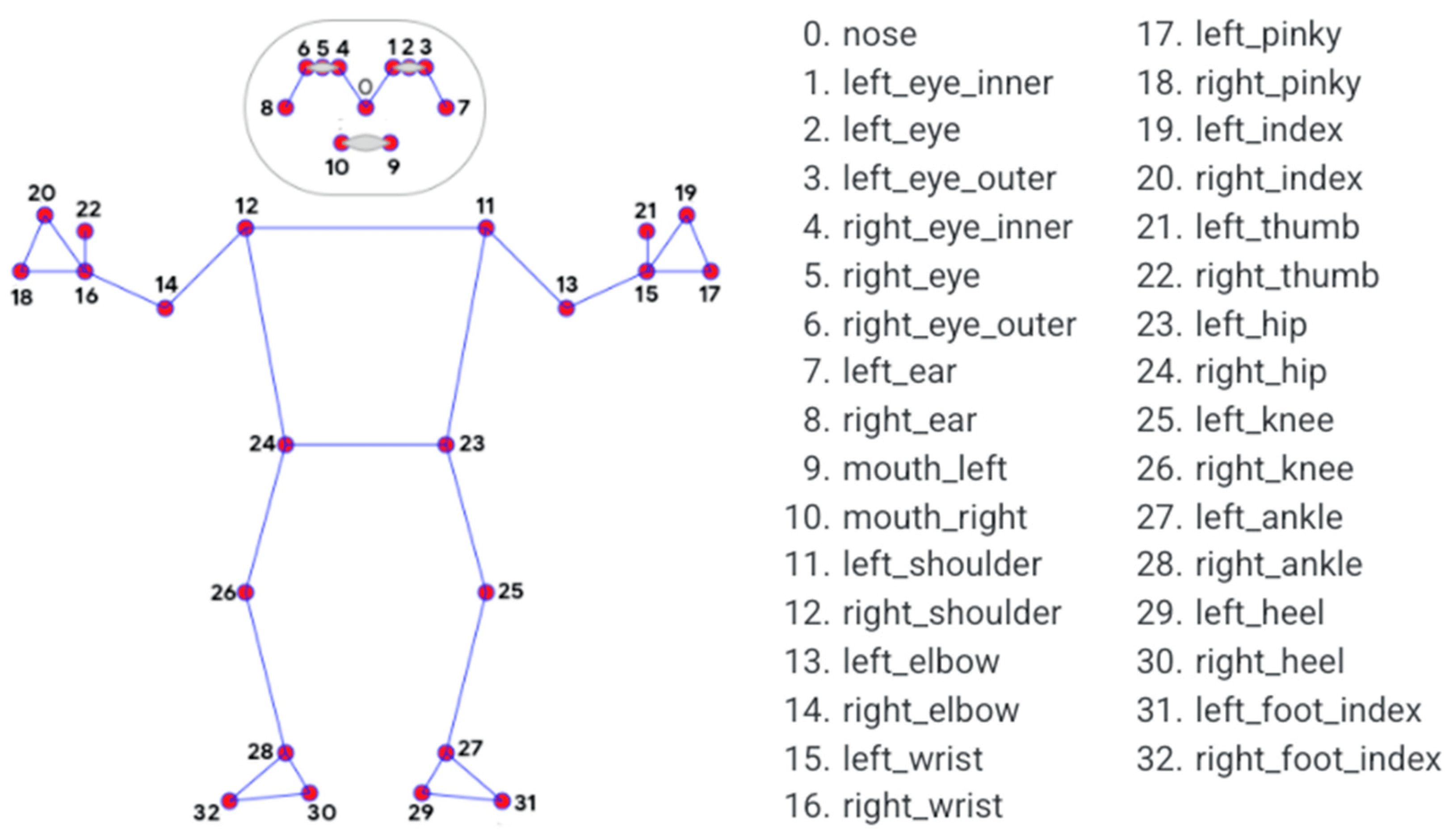} 
\caption{MediaPipe Framework Pose Estimation Joints. Source: MediaPipe GitHub page, licensed under Apache License 2.0.
Link: https://github.com/google-ai-edge/mediapipe/blob/master/docs/solutions/pose.md}
\label{mediapipe-joints}
\end{minipage}
\end{figure}

The jumps were captured using a proprietary smartphone application using version 0.7 of the Google MediaPipe framework, which was the latest version at the start of our data capture. Figure \ref{smartphone-app} contains an example of the interface of our application. The MediaPipe framework contains a pose estimation module that estimates the three-dimensional positions of 33 joints, as shown in Figure \ref{mediapipe-joints}. For this paper, we used the full version of the pose estimation module.
This module uses the BlazePose model in the backend, which was trained on 60,000 human-annotated images of one or more people in common poses. It utilizes the standard COCO topology and a CNN architecture to extract pose information. We use the estimated joint positions to store 25 joint positions, each represented by X, Y, and Z coordinates. Since the Z value range provided by the MediaPipe framework differed from that of the X and Y coordinates, we divided the Z values by a factor of 4. This factor was empirically determined by manual inspection of recorded sequences to ensure anatomically plausible knee angles.

Recordings were captured in real time using a Google Pixel 4 and a Google Pixel 6. As a result, the recordings had variable frame rates, ranging from 15 to 30 frames per second. To standardize the dataset and achieve smoother transitions between estimated poses, we upsampled all recordings to 60 frames per second using the Python scipy module's interp1d function. Adequate lighting conditions were ensured, along with a clear background to avoid interference. 

For the DVJ, we extract 30 consecutive frames from each recording, starting from 15 frames before initial contact with the floor after dropping from the box. We cut the recordings at 15 frames after initial contact, which was the mean duration until maximum knee flexion during landing, across all participants. For the SLH, we extract 15 consecutive frames beginning 30 frames prior to initial ground contact after the forward jump, encompassing the entire propulsion phase. This segment was selected because the propulsive forces generated during push-off influence lower-limb alignment and knee kinematics, providing movement information that may be comparable to the high-load landing phase of the DVJ. Initial contact with the floor for both athletic stress tests was extracted from the joint positions and velocities recorded during the tests.

\subsection{Model}

\begin{figure}[ht]
\begin{center}
\centerline{\includegraphics[width=0.60\textwidth]{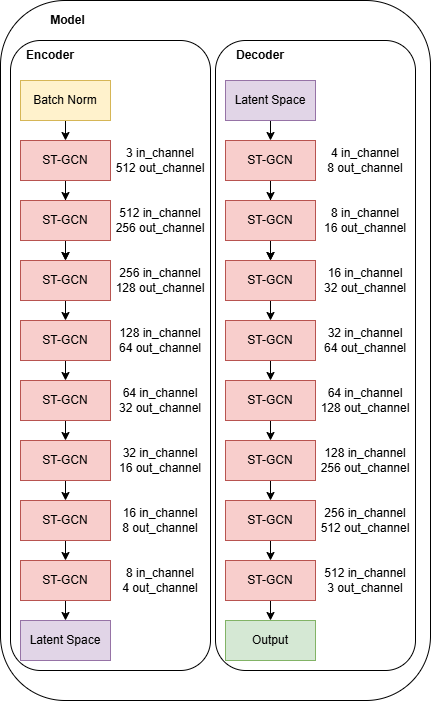}}
\caption{Model Architecture Overview}
\label{model-architecture}
\end{center}
\end{figure}

Our model uses an autoencoder architecture with multiple layers of the ST-GCN unit in the encoder and decoder. The first layer of the encoder transforms the input channels to 512, and each consecutive layer halves the number of input channels until the output channels are down to 4. The decoder performs the same operations but in reverse. This architecture is effective because the encoder initially expands the input to a much higher-dimensional feature space, allowing the capture of rich, diverse spatial-temporal patterns, before progressively compressing them into a compact latent representation that focuses on high-level patterns. Figure \ref{model-architecture} shows an overview of the model architecture.

The underlying structure of the graph convolution operation was kept identical to the original ST-GCN implementation, with the exception of the activation functions. To enhance stability with our limited dataset, we substituted ReLU with ELU, which incurs a small additional computational cost. Unlike ReLU, ELU has a smooth gradient for negative values, which reduces dying neurons and can help small datasets train more effectively. The ST-GCN model uses three different partitioning strategies for the body joints to generate a label map for the convolutions. In our experiments, we found that spatial configuration partitioning worked best. The max\_hop parameter was set to 2, and the dilation parameter to 1. These two parameters define the maximum distance between two connected nodes and the spacing between the kernel points, respectively. They are used when constructing the adjacency matrix that represents the joint structure. The temporal kernel size of each layer was set to 9, matching the original ST-GCN implementation. The stride was set to 1 on all layers. 

When determining which joints to include as input, we expected that participants' arm movements might introduce noise into the data, as they were not standardized and, as a result, might hinder the anomaly detection. To test this theory, we decided to train and compare two models: one using all joints, and another using only the joints for the lower extremities and the trunk. However, since some upper-body features still contain important information, we also compared with a third set of joints, again the joints of the lower extremities and the trunk and including the center of gravity (CoG) of the lower extremities and the trunk.  

The CoG was estimated using a weighted average of 25 joints. The corresponding joint weights used for the CoG calculation are shown in Table \ref{cog-joint-weights}, which were based on the work by Yeung et al. \cite{Yeung2014-qj}. The skeleton graph connects the CoG to the left and right hip joints. Table \ref{joint-set-metrics} shows the sensitivity, specificity and F1 score of a single cross-validation run of the three joint configurations for a DVJ, which shows that the F1 score increases quite a bit when removing the upper extremity joints but keeping the center of gravity information.

\begin{table}[ht]
\centering
\begin{tabular}{lclc}
\toprule
\textbf{Joint} & \textbf{Weight} & \textbf{Joint} & \textbf{Weight} \\
\midrule
Spine Base      & 0.213 & Right Wrist     & 0.010 \\
Spine Mid       & 0.142 & Right Hand      & 0.006 \\
Spine Shoulder  & 0.028 & Right Hand Tip  & 0.003 \\
Neck            & 0.081 & Right Thumb     & 0.003 \\
Head            & 0.069 & Left Hip        & 0.071 \\
Left Shoulder   & 0.028 & Left Knee       & 0.100 \\
Left Elbow      & 0.016 & Left Ankle      & 0.023 \\
Left Wrist      & 0.010 & Left Foot       & 0.014 \\
Left Hand       & 0.006 & Right Hip       & 0.071 \\
Left Hand Tip   & 0.003 & Right Knee      & 0.100 \\
Left Thumb      & 0.003 & Right Ankle     & 0.023 \\
Right Shoulder  & 0.028 & Right Foot      & 0.014 \\
Right Elbow     & 0.016 &                 &       \\
\bottomrule
\end{tabular}
\caption{Joint Weights for Center of Gravity Calculation}
\label{cog-joint-weights}
\end{table}

We therefore decided to only use the following joints as input to our model, along with the CoG:

\begin{itemize}
    \item \textbf{Lower Body Joints}: Left Hip, Left Knee, Left Ankle, Right Hip, Right Knee, Right Ankle
    \item \textbf{Upper Body Joints}: Spine Base, Spine Mid, Spine Shoulder
\end{itemize}

The final input to the model consists of the nine joints selected from the lower and upper body, plus the CoG, represented as a joint, for a total of 10 joints. Since we keep the X, Y, and Z positions for each joint, the size of each data point is 900 for the DVJ and 450 for the SLH. On a Google Pixel 6, the model’s average inference time was approximately one second for both jump types.

\begin{table}[ht]
\centering
\begin{tabular}{p{7cm}ccc}
\hline
\textbf{Configuration} & \textbf{Sens.} & \textbf{Spec.} & \textbf{F1} \\
\hline
All Joints & \textbf{0.88} & 0.71 & 0.62 \\
Lower Extremities + Trunk & 0.83 & 0.69 & 0.59 \\
Lower Extremities + Trunk + CoG & 0.83 & \textbf{0.81} & \textbf{0.69} \\
\hline
\end{tabular}
\caption{Performance Metrics for Different Joint Sets}
\label{joint-set-metrics}
\end{table}

\section{Experimental Results and Analysis}

\subsection{Experimental Environment}

The experiments were conducted on an Ubuntu 20.04 operating system, running on a Windows 11 laptop within the Windows Subsystem for Linux (WSL) version 2 environment. The hardware utilized for the experiments included an Intel i7-12700H processor and an NVIDIA RTX 3060 graphics card with 6GB of VRAM, which was used with CUDA version 12.1 to accelerate computation. Training was performed using the PyTorch deep learning framework version 2.4.0.

\subsection{Metrics and Parameters}

To test the performance of the model we calculated the macro-averaged F1-Score, as well as the Sensitivity and Specificity of the model, which are commonly used to assess similar screening tests. 

We randomize the order of participants and split the dataset into 90\% for training and validation and 10\% for testing. During training and validation, performance was evaluated using the Mean Squared Error (MSE) as the loss function, defined by the squared \( L_2 \) norm:
\begin{equation}
\text{Loss} = \frac{1}{N} \| \mathbf{y} - \hat{\mathbf{y}} \|_2^2
\label{eq:l2norm}
\end{equation}
In addition, the Adam optimizer was used and the learning rate was set to $3 \times 10^{-4}$ for the DVJ model and $1 \times 10^{-3}$ for the SLH model. These values were experimentally chosen to avoid both overfitting and local minimums.

We used the following strategy to evaluate the model's performance on unseen data. First, we compute the reconstruction loss for each recording. The loss values are normalized using a robust z-score based on the median and median absolute deviation (MAD) of the training loss distribution. The z-score is calculated as:

\begin{equation}
\text{loss}_z = 0.6745 \cdot 
\frac{\text{loss} - \mathrm{median}(\mathcal{L}_{\text{train}})}
{\mathrm{MAD}(\mathcal{L}_{\text{train}})}
\label{eq:robust-zscore}
\end{equation}
where \( \mathcal{L}_{\text{train}} \) represents the training loss distribution and

\begin{equation}
\mathrm{MAD}(\mathcal{L}_{\text{train}}) =
\mathrm{median}\left(
\left|
\mathcal{L}_{\text{train}} -
\mathrm{median}(\mathcal{L}_{\text{train}})
\right|
\right).
\end{equation}

The constant \(0.6745\) scales the MAD so that the resulting value is comparable to a standard z-score under the assumption of normally distributed data. To classify recordings as high or low risk, we define a threshold based on the distribution of z-scores computed during validation:

\begin{equation}
t = Q_{1-\alpha}(\text{loss}_z)
\end{equation}
where \(Q_{1-\alpha}\) denotes the \((1-\alpha)\)-quantile of the z-score distribution and \(\alpha\) represents the target false positive rate. We used \(\alpha = 0.25\) for both models and recordings with \( \text{loss}_z > t \) were classified as high risk.

For each athletic stress test, the training and validation data were divided into five folds and cross-validation was performed using an 80/20 split, for a fixed test set. In each fold, a separate model was trained for five epochs with a batch size of 32. To avoid data leakage issues, we ensure that the training, validation and test sets are individual-wise mutually exclusive.

To better estimate real-world performance, this 5-fold cross-validation procedure was repeated five times with different test sets. In each repetition, the test set consisted of a different subset of participants, 10\% selected sequentially from the entire data set, while the remaining participants were used for training and validation. Performance metrics were then averaged across all folds and repetitions, comparing 25 models against 5 test sets in total. This evaluation protocol is similar to previously proposed methods for performance estimation \cite{Jung2015-ks}.

\subsection{Results}

Table \ref{model-performance} contains the average metrics on the test set for each of the five cross-validation runs, as well as the averages across all runs for the DVJ and the SLH. The specificity can differ quite a bit from run to run, which may be due to individuals included in the test set with poor biomechanics who avoided injury during the follow-up period. Furthermore, the F1 scores are relatively low, especially for the SLH, which is expected given the highly unbalanced dataset and the possibility that more high-risk athletes who avoided injury during the follow-up time are included.

\begin{table}[ht!]
\centering
\begin{tabular}{c|ccc|ccc}
\toprule
\textbf{Run} & \multicolumn{3}{c|}{\textbf{Drop Vertical Jump (DVJ)}} & \multicolumn{3}{c}{\textbf{Single Leg Hop (SLH)}} \\
            & \textbf{Sensitivity} & \textbf{Specificity} & \textbf{F1 Score} 
            & \textbf{Sensitivity} & \textbf{Specificity} & \textbf{F1 Score} \\
\midrule
1 & 83\% & 76\% & 0.65 & 76\% & 70\% & 0.50 \\
2 & 83\% & 81\% & 0.69 & 76\% & 77\% & 0.55 \\
3 & 71\% & 83\% & 0.68 & 76\% & 75\% & 0.53 \\
4 & 76\% & 82\% & 0.68 & 73\% & 74\% & 0.52 \\
5 & 78\% & 68\% & 0.58 & 73\% & 76\% & 0.53 \\
\midrule
\textbf{Mean} & \textbf{78\%} & \textbf{78\%} & \textbf{0.66} 
             & \textbf{75\%} & \textbf{74\%} & \textbf{0.53} \\
\bottomrule
\end{tabular}
\caption{Repeated Cross-Validation Results for DVJ and SLH (5 Runs) \label{model-performance}}
\end{table}

Table \ref{performance-comparison} contains a comparison of our DVJ model with previous attempts to predict the risk of ACL injury using the same test. An important note is that the method by Zazulak et al.~used a specialized core proprioception measurement apparatus, the method by Bates et al.~used data from a complete motion capture lab setup, and Corban et al.~used a desktop application that works with the Kinect V2 sensor \cite{Zazulak2007-gw, Bates2020-tb, Corban2023-ns}. Our model can run on a single smartphone in real time and work in any well-lit environment. A similar table for the SLH is not provided, as, to the best of our knowledge, it has not been used before for ACL injury risk prediction.

\begin{table}[ht]
\centering
\begin{tabular}{ccc}
\toprule
\textbf{Method} & \textbf{Sensitivity} & \textbf{Specificity} \\ \midrule
Zazulak et al. \cite{Zazulak2007-gw} & 0.90 & 0.56 \\ 
Bates et al. \cite{Bates2020-tb} & 0.78 & 0.83 \\ 
Corban et al. \cite{Corban2023-ns} & 0.80 & 0.74 \\ 
Ours & 0.78 & 0.78 \\ 
\bottomrule
\end{tabular}
\caption{Sensitivity and Specificity of Injury Prediction Using the DVJ\label{performance-comparison}}
\end{table}

\begin{figure}[ht]
\begin{minipage}{0.48\textwidth}
\centering
\includegraphics[width=0.99\columnwidth]{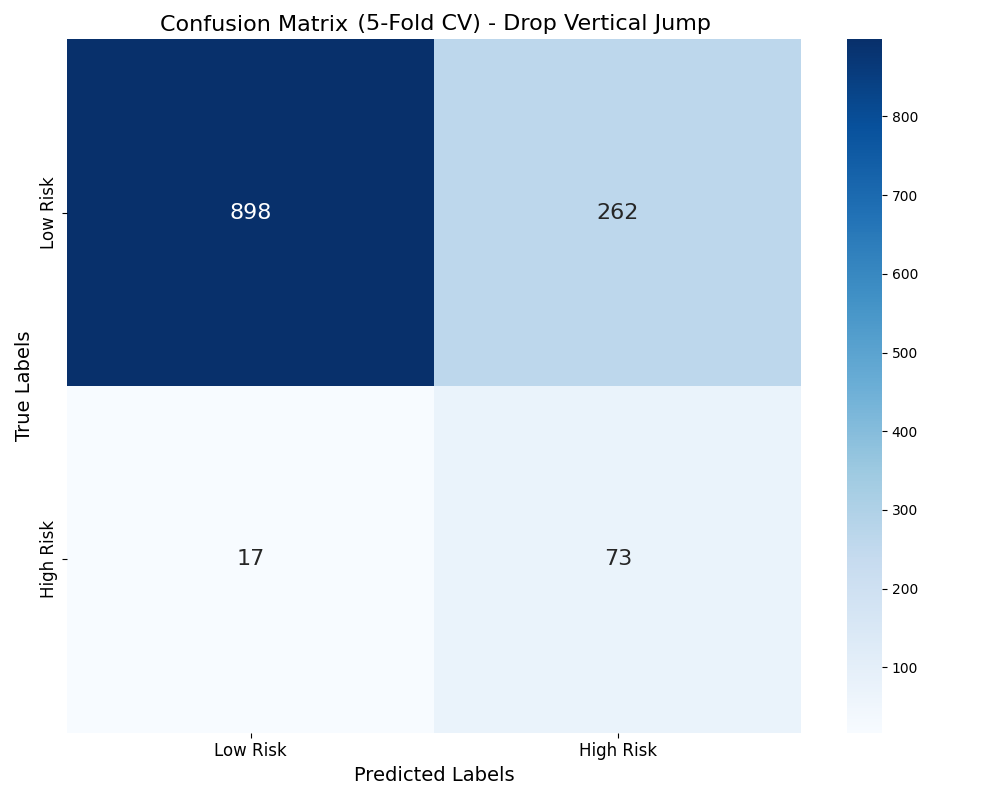} 
\caption{Confusion Matrix (5-Fold CV) - Drop Vertical Jump}
\label{confusion-matrix-heatmap-dvj}
\end{minipage}\hfill
\begin{minipage}{0.48\textwidth}
\centering
\includegraphics[width=0.99\columnwidth]{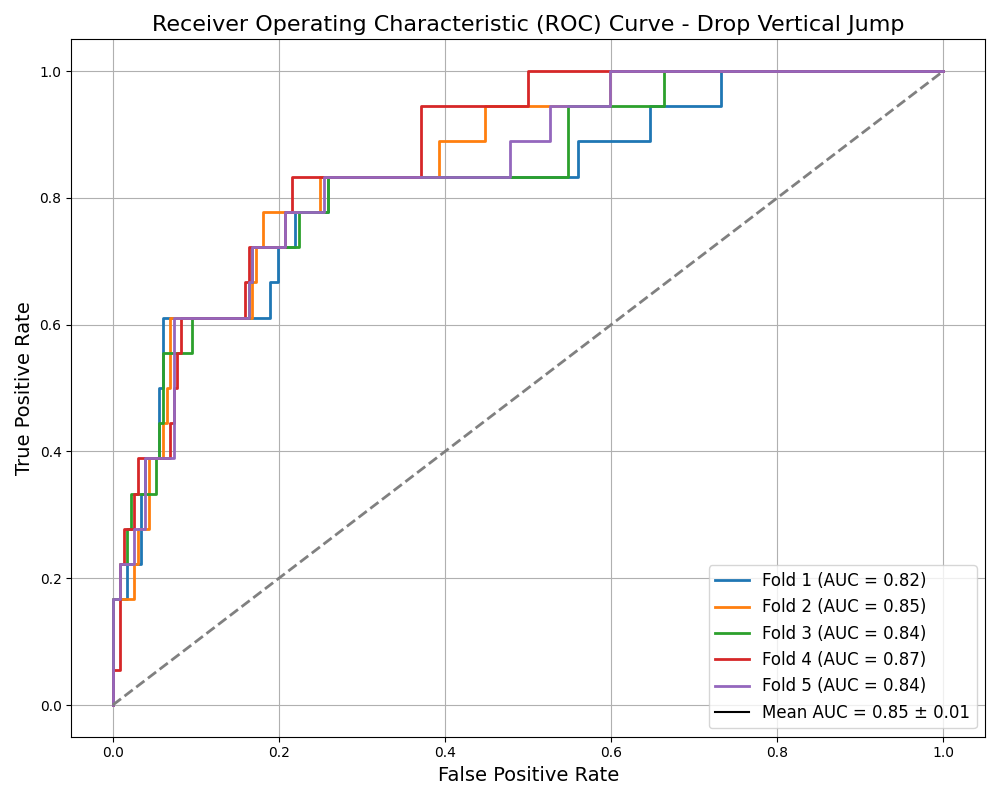} 
\caption{Receiver Operator Characteristic (ROC) Curve (5-Fold CV) - Drop Vertical Jump}
\label{roc-curve-dvj}
\end{minipage}
\end{figure}

Figure \ref{confusion-matrix-heatmap-dvj} contains a confusion matrix taken from the first cross-validation run (run 1) for the DVJ, generated from the aggregated labels and z-scores of the five folds. As shown in the figure, a high negative predictive value (98.1\%) but a low positive predictive value (21.8\%) is observed; the sensitivity of the model should be interpreted cautiously, as there were very few positive cases. 

The receiver operator characteristic (ROC) area under the curve (AUC) across all runs was 0.83 \(\pm\) 0.04 for the DVJ, indicating strong discriminative performance. Figure \ref{roc-curve-dvj} shows the ROC curves from one representative cross-validation run (run 1) of the DVJ, for illustrative purposes. The AUC is consistent across folds, suggesting that the model performs similarly when trained on different subsets of the data.

\begin{figure}[ht]
\begin{minipage}{0.48\textwidth}
\centering
\includegraphics[width=0.99\columnwidth]{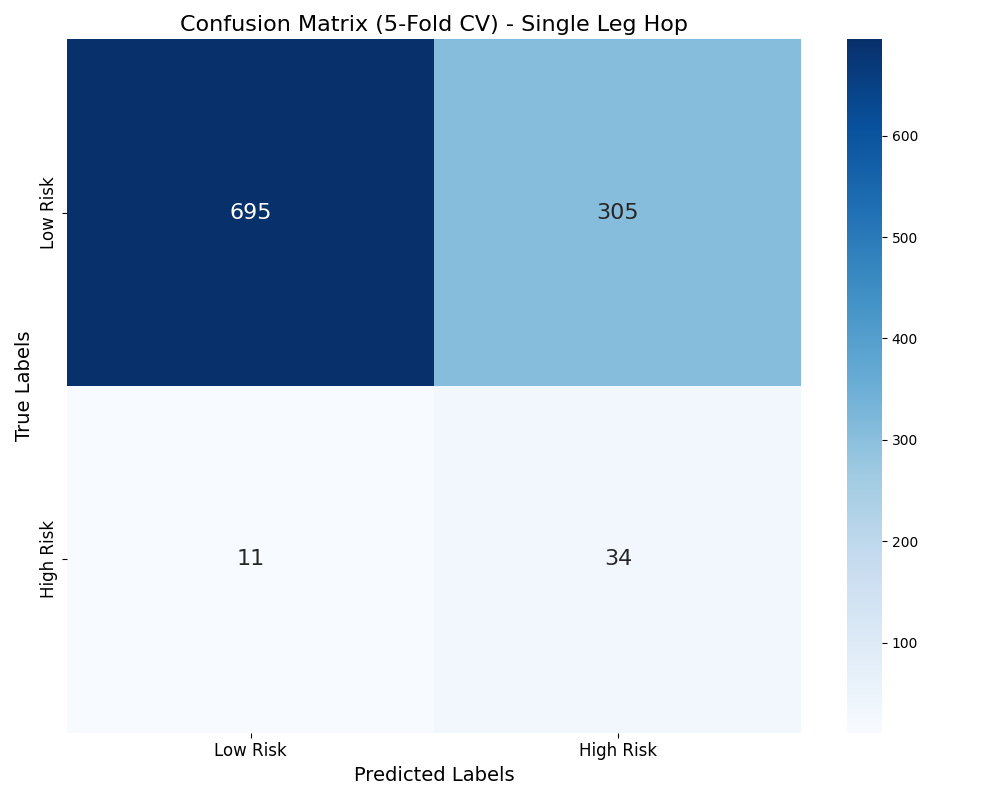} 
\caption{Confusion Matrix (5-Fold CV) - Single Leg Hop}
\label{confusion-matrix-heatmap-slh}
\end{minipage}\hfill
\begin{minipage}{0.48\textwidth}
\centering
\includegraphics[width=0.99\columnwidth]{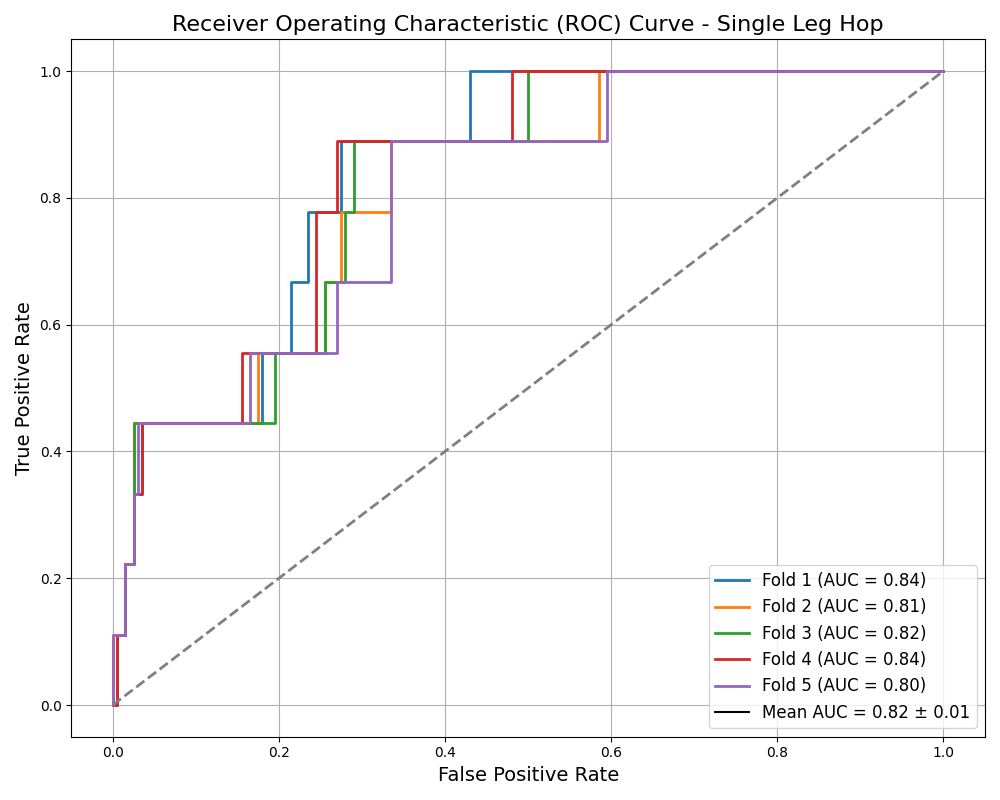} 
\caption{Receiver Operator Characteristic (ROC) Curve (5-Fold CV) - Single Leg Hop}
\label{roc-curve-slh}
\end{minipage}
\end{figure}

Similarly, Figures \ref{confusion-matrix-heatmap-slh} and \ref{roc-curve-slh} contain a confusion matrix and the ROC curves from one representative cross-validation run (run 1) for the SLH, respectively. The ROC AUC for this athletic stress test across all runs was 0.85 \(\pm\) 0.01, also indicating strong discriminative performance. The trends in the figures resemble those of the DVJ, although overall performance is slightly lower, with modest reductions in accuracy and positive predictive value.

\begin{figure}[ht]
\centering
\includegraphics[scale=0.42]{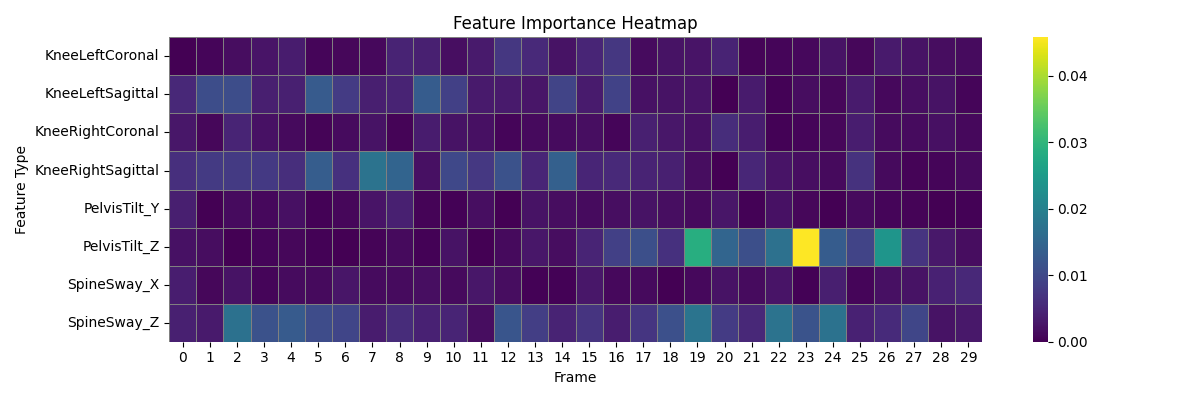}
\caption{Drop Vertical Jump Angle Importance Heatmap}\label{dvj-angle-importance}
\end{figure}

\begin{figure}[ht]
\centering
\includegraphics[scale=0.42]{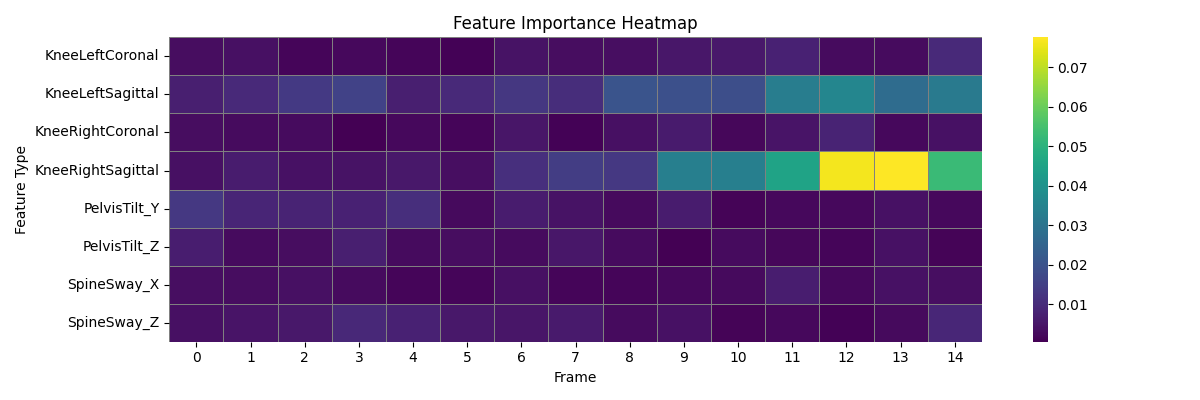}
\caption{Single Leg Hop Angle Importance Heatmap}\label{slh-angle-importance}
\end{figure}

Finally, Figures \ref{dvj-angle-importance} and \ref{slh-angle-importance} show two heatmaps of the importance of certain key joint angles during a DVJ and an SLH, respectively, over the extracted frame section. These were generated from the predictions of the model on the test set of the first cross-validation run. Table \ref{angle_variables} contains an explanation of the variables in the figures. For the DVJ, an important factor seems to be the pelvic rotation after initial contact with the floor, as shown in Figure \ref{dvj-angle-importance} at frame 15. This elevated pelvic rotation might indicate an uneven landing pattern, leading to increased stress on one knee. For the SLH, the sagittal knee angle just before leaving the ground during the propulsion phase is identified as a key contributor. Reduced ground reaction forces during propulsion have been associated with increased injury risk, a phenomenon the model likely captures indirectly through learned movement patterns \cite{Nardon2024-wq}.

\begin{table}[ht]
\small
\centering
\caption{Descriptions of Biomechanical Parameters Used for Figures \ref{dvj-angle-importance} and \ref{slh-angle-importance}}
\begin{tabular}{|l|l|}
\hline
\textbf{Variable} & \textbf{Explanation} \\
\hline
\texttt{KneeLeftCoronal} & Angle of the left knee in the coronal plane. \\
\hline
\texttt{KneeLeftSagittal} & Angle of the left knee in the sagittal plane. \\
\hline
\texttt{KneeRightCoronal} & Angle of the right knee in the coronal plane. \\
\hline
\texttt{KneeRightSagittal} & Angle of the right knee in the sagittal plane. \\
\hline
\texttt{PelvisTilt\_Y} & Angle measuring pelvic obliquity. \\
\hline
\texttt{PelvisTilt\_Z} & Angle measuring pelvic rotation. \\
\hline
\texttt{SpineSway\_X} & Lateral sway of the spine. \\
\hline
\texttt{SpineSway\_Z} & Anterior-posterior sway of the spine. \\
\hline
\end{tabular}
\label{angle_variables}
\end{table}

\subsection{Discussion and Limitations}

Our findings suggest that movement patterns associated with ACL strain can be accurately identified using only real-time data captured from standard mobile devices. Although these results are preliminary because of the limited and imbalanced dataset, they provide important proof of concept that pose estimation data can reveal motion characteristics linked to soft tissue injury risk. The ROC AUC values indicate that both developed models demonstrate strong discriminative performance. With a larger, more balanced dataset, optimization of the decision threshold would likely further improve overall model performance.

One of the primary concerns with our dataset is that the positive cases are not clearly defined. Despite following up with athletes for more than a year, some athletes in the cohort may have avoided injury during this period but are still assessed as at risk of ACL injury. In our initial attempts to develop a machine learning model for this purpose, we experimented with multiple supervised learning models to classify athletes, but all performed poorly. We hypothesized that several participants in the cohort may be false negatives, which could have contributed to challenges during model training. As such, we decided to use an unsupervised learning approach, which alleviates this concern by focusing on learning the high-risk pattern without supervision.

Although this approach reduces the influence of false negatives on the model weights, there are still two considerations to keep in mind. Firstly, since our dataset is relatively small, if the training subset contains many individuals at high risk of injury, the model may misclassify individuals with similar biomechanics as low risk. Secondly, when testing the model on participants who did not sustain an injury during the follow-up period but exhibit high-risk biomechanics, the model may still misclassify them as false positives. For these reasons, we performed the training and testing multiple times and averaged the results to better represent the performance.

Our dataset consists exclusively of varsity athletes, who are generally highly trained and expected to perform the exercise proficiently. This feature makes the dataset particularly well-suited for capturing clean and consistent movement patterns. However, because the participants represent a specialized population, the model may not generalize fully to the broader public. Additionally, although we included a substantial number of athletes, the dataset remains relatively small by deep learning standards. Future work involving a larger, more diverse cohort would help validate and potentially extend the model's applicability.

\section{Conclusion}

In this paper, we propose a deep learning model to predict the risk of ACL injury by analyzing an individual's motion during a DVJ or SLH. The model is trained on a proprietary dataset of pose estimation results, captured in real time on a modern smartphone using a widely available framework. Our results are comparable to previous approaches, which relied primarily on calculating joint angles at specific key points during the athletic stress test and used expensive or currently unavailable technologies. By contrast, this work demonstrates that modern machine learning methods can identify movement patterns associated with ACL strain using widely accessible technology. Reducing the need for costly or laboratory-based systems represents an important step toward democratizing injury prevention, enabling real-time movement assessment on standard mobile devices and expanding access beyond specialized clinical or research settings.

\section*{Acknowledgements}
The second author would like to acknowledge the support of Semperform Innovation Inc. to this research endeavour.

\bibliographystyle{splncs04}
\bibliography{main}
\end{document}